\documentclass[letterpaper]{article}

\usepackage{natbib,alifeconf}  
\usepackage{url,hyperref,formatting}
\usepackage{enumitem}
\usepackage{cleveref}
\usepackage{booktabs}
\usepackage{nameref}

\title{Guiding Evolution of Artificial Life Using Vision-Language Models}

\author{
   Nikhil Baid$^{*1}$,
   Hannah Erlebach$^{*1}$,
   Paul Hellegouarch$^{*1,2}$, \and
   Frederico Wieser$^{*1}$ \\
   \mbox{}\\
   * \textit{Contributed equally}\\
   $^1$University College London, UK\\
   $^2$ Institut Pasteur, Université Paris Cité, CNRS UMR 3571, Decision and Bayesian Computation, Paris, France \\
   nikhil.baid.24@ucl.ac.uk, hannah.erlebach.24@ucl.ac.uk \\
   paul.hellegouarch@pasteur.fr, fred.wieser.18@ucl.ac.uk
} 

\begin{document}

\maketitle

\begin{abstract}

Foundation models (FMs) have recently opened up new frontiers in the field of artificial life (ALife) by providing powerful tools to automate search through ALife simulations. Previous work aligns ALife simulations with natural language target prompts using vision-language models (VLMs). We build on Automated Search for Artificial Life (ASAL) by introducing \MethodName{}, a method for open-ended-like search guided by multimodal FMs. We use a second FM to propose new evolutionary targets based on a simulation’s visual history. This induces an evolutionary trajectory with increasingly complex targets. We explore two strategies: (1) evolving a simulation to match a single new prompt at each iteration (\textit{Evolved Supervised Targets}: EST) and (2) evolving a simulation to match the entire sequence of generated prompts (\textit{Evolved Temporal Targets}: ETT). We test our method empirically in the Lenia substrate using Gemma-3 to propose evolutionary targets, and show that EST promotes greater visual novelty, while ETT fosters more coherent and interpretable evolutionary sequences. Our results suggest that \MethodName{} points towards new directions for FM-driven ALife discovery with open-ended characteristics.

\end{abstract}



\textbf{Code and examples are available at}:

https://github.com/fredericowieser/ASALPlusPlus
\section{Introduction}
\label{sec:Introduction}
Cellular automata were introduced by Stanislaw Ulam and John von Neumann in the 1940s \citep{vonNeumann1966Theory}, demonstrating how discrete rules could lead to complex emergent behaviours. 

A classic example is Conway's Game of Life (GoL) \citep{gardner1970fantastic}, which exhibits chaotic patterns starting from simple rules in a discrete space. In 1987, Christopher Langton brought together multiple disciplines to coin the field of artificial life (ALife) \citep{langton1989artificial}, with the aim of studying life not just as it is, but as it could be. Cellular automata such as GoL have been integral to the study of virtual ALife systems. A particular area of interest is \textit{open-ended evolution}---creating evolving systems which produce unbounded complexity and never settle into a single stable equilibrium \citep{ruiz2004universal, soros2014identifying}.

\begin{figure}[H]
    \centering
    \includegraphics[width=0.95\linewidth]{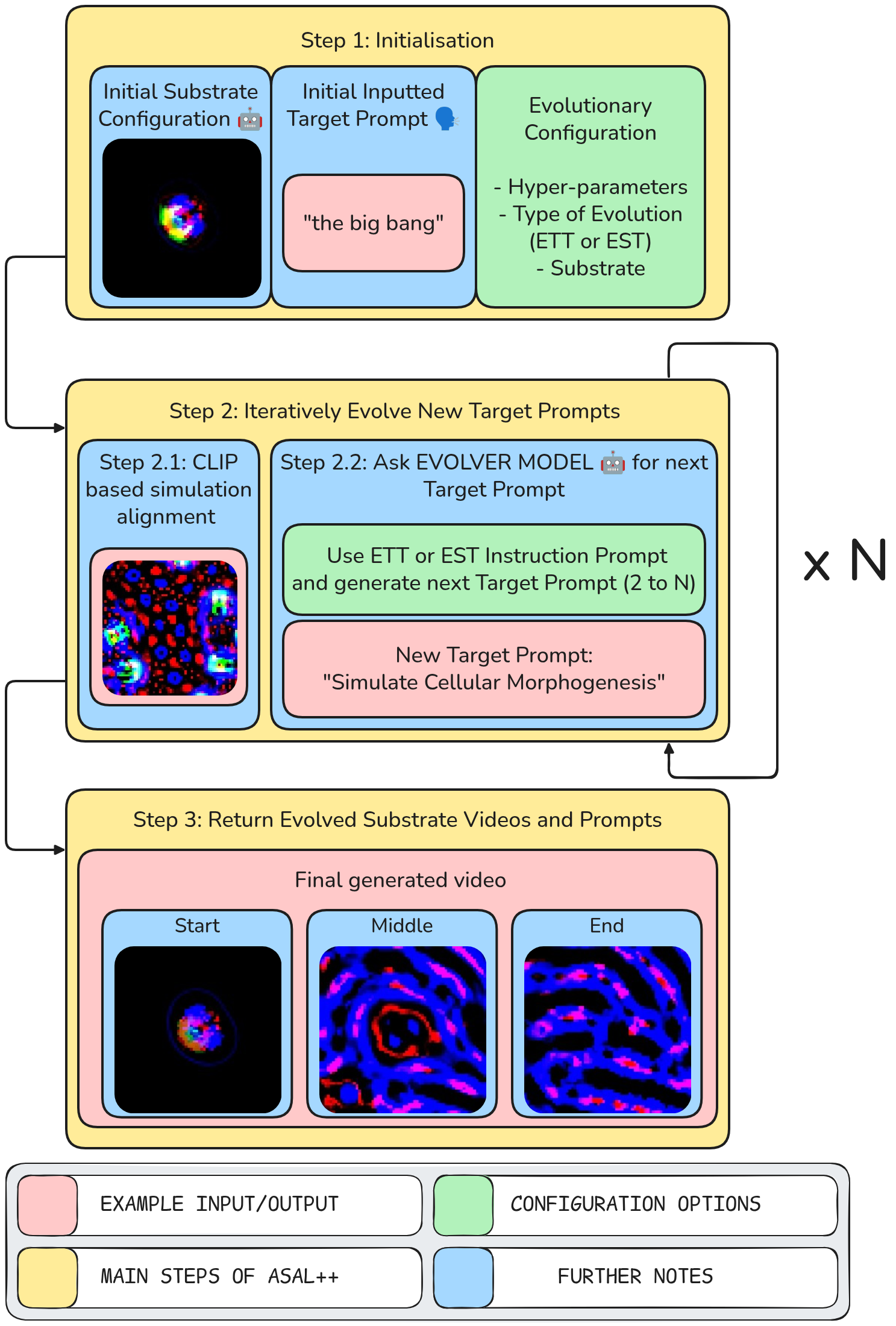}
    \caption{\textbf{Schematic of \MethodName{}}. A seed prompt initiates a loop that alternates between CLIP-based simulation alignment and autonomous prompt generation by a second model, guiding the simulation's emergent evolution.}
    \label{fig:intro-diagram}
\end{figure}

Foundation models (FMs) \citep{bommasani2021opportunities} have recently created new possibilities for open-endedness \citep{hughes2024position}, including paving the way for novel search methods in ALife. This has been demonstrated in the open-ended evolution of prompts \citep{fernando2024promptbreeder}, generation of open-ended tasks \citep{faldor2024omni} or fully AI-automated scientific research \citep{lu2024ai}. In addition, their latent representation space can capture meaningful relationships between inputs \citep{huh2024platonic}. Our aim in this work is to \textbf{leverage foundation models as drivers of search through artificial life simulations}, to discover continually novel trajectories.

Our work builds on the method introduced by \citet{kumar2024automating}; Automating the Search for Artificial Life (ASAL). 
ASAL uses FMs---specifically multimodal models with shared latent spaces for images and text, like Contrastive Language-Image Pre-training (CLIP) \citep{radford2021learningtransferablevisualmodels} ---to guide ALife simulations towards user-defined target prompts by optimising the simulation parameters based on the visual and textual embeddings.

Our method, \MethodName{}, introduces a second FM, 
the \EvolverModel{}, prompted to autonomously generate novel targets based on historical trajectories. Each iteration of our method runs ASAL as an ``inner loop'' on targets proposed by the \EvolverModel{}. This automatically guides trajectories in which the \EvolverModel{} proposes ever more complex targets. \MethodName{} is illustrated in \Cref{fig:intro-diagram}.



We use Gemma-3, specifically the recent multi-modal FM \texttt{gemma-3-4b-it}, as the \EvolverModel{} in \MethodName{}. We focus on Lenia \citep{wang2019lenia}, a popular ALife substrate which is a continuous generalisation of Smooth Life.  We either use a single target or temporal targets that consider multiple Gemma-3 outputs (\Cref{fig:intro-diagram}), both of which result in interesting open-ended behaviours. We also prompt Gemma-3 more than once at each iteration to produce tree structures that show the exploratory capability of our method. Lastly, we find limitations in our method from the FM, such as converging to local minima and computational cost. 




To our knowledge, our work is the first to apply ideas from open-endedness for a \textbf{foundation model-driven search} in the evolution of ALife. 


\section{Background}
\label{sec:Background}
\subsection{Artificial life}

The study of ALife focuses on understanding, replicating and extending the fundamental principles of life, with the support of \textbf{substrates}. GoL is an example of a two-dimensional substrate where its evolution is entirely determined by its initial configuration.

We follow the formulation in \citet{kumar2024automating} to describe a substrate. A \text{substrate} $S$ refers to a parametrised family of ALife simulations, each defined by a specific set of parameters $\theta$. A given $\theta$ defines a single simulation instance via the three following  components:

\begin{enumerate}
    \item An initial state distribution: $s_0 \sim \text{Init}_\theta$
    \item A forward dynamics step function: $s_{t+1} = \text{Step}_\theta(s_t)$
    \item A rendering function $\text{Render}_\theta$ which produces an image from a state
\end{enumerate}

By running these simulations for $T$ time steps, we define a simulated trajectory rendering $RS^{T}(\theta)$ as follows:
\begin{equation}
RS^{T}(\theta) = \text{Render}_\theta(\text{Step}_\theta^T(s_0))
\end{equation}
where $\text{Step}_\theta^T(s_0)$ denotes the result of applying the step function $\text{Step}_\theta$ iteratively for $T$ timesteps starting from the initial state $s_0$. We refer to a simulated trajectory as a \textit{rollout}. 

\subsection{Open-endedness}

\textit{Open-endedness}---known in the ALife community as \textit{open-ended evolution}---is a field in AI which aims to find algorithms that can in principle discover novel, interesting artefacts indefinitely \citep{hughes2024position}. A key element of open-ended search is \textbf{stepping stones} \citep{stanley2015greatness}: intermediate discoveries which precede more interesting and valuable discoveries, without the foresight of how they will be used. Recently, FMs have increasingly been used as drivers of open-ended search, given their ability to represent the world as well as human notions of interestingness \citep{zhang2024omniopenendednessmodelshuman}. Works such as the AI Scientist \citep{lu2024ai} showcase the burgeoning ability of FMs to generate novel, creative proposals. This paves the way for research into uses of these FMs to automate the search for novel, interesting artifacts in diverse domains.

\subsection{Automating the Search for Artificial Life}

ASAL uses latent FM representations to drive the search for ALife simulations. In particular, ASAL uses \textbf{vision-language models} (VLMs) equipped with embedding functions which map images and text into an embedding space, allowing for metrics such as cosine similarity to be applied:

\vspace{-0.3em}

$$
\text{VLM}_{\text{img}}(\cdot) \quad \text{and} \quad \text{VLM}_{\text{txt}}(\cdot)
$$

\vspace{-0.3em}

Although various VLMs can be chosen, ASAL primarily uses CLIP, a vision-language FM trained on text-image pairs which explicitly provides embedding functions. 

ASAL implements a number of objectives using these embeddings, and we focus on the \textit{supervised targets}.


\subsubsection{Supervised targets}

The objective is to find simulation parameters $\theta$ whose sequence of frames is semantically close to given target natural language prompts in FM embedding space. For a \textit{single target} prompt, the optimisation criterion is:
\begin{multline}
\theta^* = \arg\max_\theta \, \mathbb{E}_T \Big\langle 
\text{VLM}_{\text{img}}\big(RS^{T}(\theta)\big), \\
\text{VLM}_{\text{txt}}(\text{prompt}_T)
\Big\rangle
\label{eq:supervised_target}
\end{multline}

\Cref{eq:supervised_target} can also be used to find simulations producing a \textbf{target sequence} of events, by optimising a list of prompts applied at evenly spaced time intervals of the simulation rollout. This is referred to as \textit{temporal targets}, and can be further amplified by introducing a bidirectional softmax similarity to promote one-to-one matching between each prompt and the corresponding section of the rollout. The relative weighting of this in the loss is controlled via the \textit{softmax coefficient}.


ASAL uses CLIP as the embedding FM and Sep-CMA-ES \citep{ros2008simple} for the optimisation algorithm.


\section{Methods}
\label{sec:Methods}
We introduce \MethodName{}, a method using FMs to dynamically generate \textbf{new supervised target prompts} for ALife simulations. Rather than relying on fixed targets, our approach encourages open-ended characteristics in trajectory generation by using FMs to dynamically propose new targets based on the simulation's evolutionary history. This enables open-ended-like discovery of increasingly complex novel behaviours.


\subsection{Evolving targets using FMs}

In ASAL \citep{kumar2024automating}, optimisation to search for targeted simulations can be performed either with a \textit{single target} (a single prompt applied at the end of the rollout) or with \textit{temporal targets} (a sequence of prompts applied at evenly spaced intervals). We previously referred to this general setup as \textbf{supervised targets}, since prompts are pre-specified and used to supervise the simulation. However, pre-selecting these prompts in advance precludes the opportunity to notice  \textit{stepping stones} during the evolutionary trajectory of the simulation. We address this by prompting an \EvolverModel{}---a FM which accepts both video and text inputs---to propose a \textbf{next target prompt}. 

In the first iteration, we run ASAL with a human-generated initial prompt. Thereafter, in each iteration of \MethodName{}, we pass a video of the current optimised simulation and in ETT's case also the current list of prompts to the \EvolverModel{}.
The \EvolverModel{} then generates the next target prompt. The new prompt is appended to the current list of prompts, and the simulation is optimised to match this list of prompts, according to \Cref{eq:supervised_target}. Importantly, we use warm starts: the initial parameters for each iteration are the best parameters found by the previous iteration. We run \MethodName{} for $N$ iterations, with $I$ inner iterations of ASAL, resulting in a generated sequence of $N$ prompts.





We introduce two versions of this method:

\begin{enumerate}[itemsep=0pt, topsep=1pt, leftmargin=*]
    \item \textbf{Evolved Supervised Targets (EST).} At each iteration we generate a new target prompt using \EvolverModel{}, and \textit{only} optimise the simulation for that single prompt. We pass the simulation video from the previous iteration and its target prompt to the \EvolverModel{}.
    \item \textbf{Evolved Temporal Targets (ETT).} At each iteration, we optimise for the \textit{entire list of prompts} generated so far. We pass the simulation video from the previous iteration and the current list of target prompts to the \EvolverModel{}. The  full algorithm for ETT is shown in \Cref{alg:evolutionary-prompting}.
\end{enumerate}

 Importantly, \MethodName{} is substrate-agnostic for any substrate with 2D video rendering, and the \EvolverModel{} can be any FM which accepts video and text inputs. 

\begin{minipage}{0.95\linewidth}
\begin{algorithm}[H]
\caption{\MethodName{} with evolved temporal targets}
\label{alg:evolutionary-prompting}
\begin{algorithmic}[1]
\Require Substrate \( S \), foundation model \( \mathcal{F} \), encoder \( \mathcal{E} \), initial target prompt \( p \), ES population size \( P \), ES step size \( \sigma \), inner iterations \( I \),  rollout steps \( T \), \MethodName{} iterations \(N\)
\Ensure Optimised parameters \( \theta^* \) minimizing loss

\State \( z_p \gets \mathcal{E}(p) \) \Comment{Embed initial prompt}
\State Initialise CMA-ES optimiser with population size \( P \) and step size \( \sigma \)

\For{each iteration \( n = 1, \dots, N \)}
    \For{each iteration \( i = 1, \dots, I \)}
        \State \( \Theta \gets \text{CMA-ES.ask()} \) 
        \For{each \( \theta \in \Theta \)}
            \State \( \text{Imgs}_{0:T} \gets \text{Simulate}_T(S, \theta) \) 
            \State \( z_{0:T} = \mathcal{E}(\text{Imgs}_{0:T})\)
            
            \State \( \mathcal{L}(\theta) \gets - \langle z_{0:T}, z_p \rangle \)

        \EndFor
        \State CMA-ES.tell(\( \Theta, \mathcal{L}(\Theta) \)) 
        \State Track \( \theta^* = \arg\min_{\theta} \mathcal{L}(\theta) \) 
    \EndFor
    \State \( p' \gets \mathcal{F}_{\text{new prompt}}(\text{Imgs}_{0:T}) \)
    \State \( p \gets p + p'\) 
\EndFor

\State \Return \( \theta^* \)

\end{algorithmic}
\end{algorithm}
\end{minipage}








\section{Experiments}
\label{sec:Experiments}
\subsection{Substrates}

\paragraph{Lenia.} 


We choose Lenia  as our primary substrate because of its expressivity and speed on limited computational resources.
Lenia is a generalisation of GoL to continuous space, time and state values, allowing for
more intricate patterns and dynamics. These patterns often display
``lifelike'' behaviours such as self-organization.




\subsection{Evolving prompts in Lenia using Gemma-3}
\label{sec:main-experiments}

We run both the EST and ETT variations of \MethodName{} in Lenia for 9 different human-chosen initial prompts. We use \texttt{gemma-3-4b-it}, a recent lightweight open-source multimodal model, as the \EvolverModel{} to propose new targets. We use 8 iterations of \MethodName{}---generating 8 sequential target prompts---and each iteration runs 2000 ASAL optimisation steps. Each rollout has the default 256 steps, meaning that we generate 256 frames for each discovered simulation. 


\Cref{fig:instruction_prompt} shows the instruction prompt received by Gemma-3 for ETT. 

\begin{figure}
\begin{ettpromptbox}
This artificial life simulation has been optimised to follow this sequence of prompts:
`\var{all\_prompts}'.

Consider these as constraints: ecological niches that have already been explored.\\

You are in iteration \var{i}.  Your task is to propose the NEXT TARGET PROMPT to determine the next stage of evolution.  This is an opportunity to propose a direction that is significantly different from the past, but leads to interesting lifelike behaviour.  Can we recreate open-ended evolution of life?  Be bold and creative!  ONLY output the new target prompt string, and be concise. Avoid using too many adjectives.\\

NEXT TARGET PROMPT:
\end{ettpromptbox}
\caption{\textbf{Instruction prompt for Gemma-3.} Gemma-3 is also passed the video of the rollout from the previous iteration of \MethodName{}.}
\label{fig:instruction_prompt}
\end{figure}


\subsection{Phylogenetic trees}

In addition, we generate a collection of alternative evolutionary trajectories by prompting Gemma-3 to output multiple new target prompts at each iteration, which we visualise in a ``tree of life'' structure. This is motivated by phylogenetic trees \citep{hennig1966phylogenetic}  in an effort to exploit the entropy of our method for open-ended evolution. We use a higher temperature regime, T=1, to encourage Gemma-3 to produce different prompts.  If it is not able to produce a new prompt after 10 tries we move on.  This can lead to unsymmetrical splitting. 

We use ETT, taking into account only the prompts that directly lead to the root. 

In an effort to encourage further diversity we also introduce environmental inputs. This method takes further inspiration from natural evolution where an organism develops as a function of its environment. We lower the temperature for this to be consistent with the main experiments and use another foundation model, GPT-4o \citep{achiam2023gpt}, to suggest environmental inputs and append these to Gemma-3 outputs.

\subsection{Open-endedness score}

We use an open-endedness (OE) score, inspired by \citet{kumar2024automating}, to measure the open-endedness of a simulation. Specifically, we compare the similarity of each frame at time step $T$ with earlier frames $T' < T$, using CLIP embedding space as our VLM. The final OE score is one minus this maximum similarity:
\begin{equation}
1 - \max_{T'<T} \Big\langle 
\text{VLM}_{\text{img}}\big(RS^{T}(\theta)\big),
\text{VLM}_{\text{img}}\big(RS^{T'}(\theta)\big)
\Big\rangle
\label{eq:oe_score}
\end{equation}
A higher score indicates greater visual novelty at time $T$ compared to all previous timesteps, and thus reflects more open-ended behaviour.



\subsection{Baselines and ablations}
For open-endedness within each iteration we use ASAL as a baseline by comparing the OE score of the first iteration which is just \textbf{supervised targets} ASAL and then see how this compares to the final iteration to see how much the novelty of individual iterations is affected through this method.

\section{Results}
\label{sec:Results and discussion}


\subsection{Evolved Supervised Target}

\begin{figure} 
    \centering
    \includegraphics[width=0.76\linewidth]{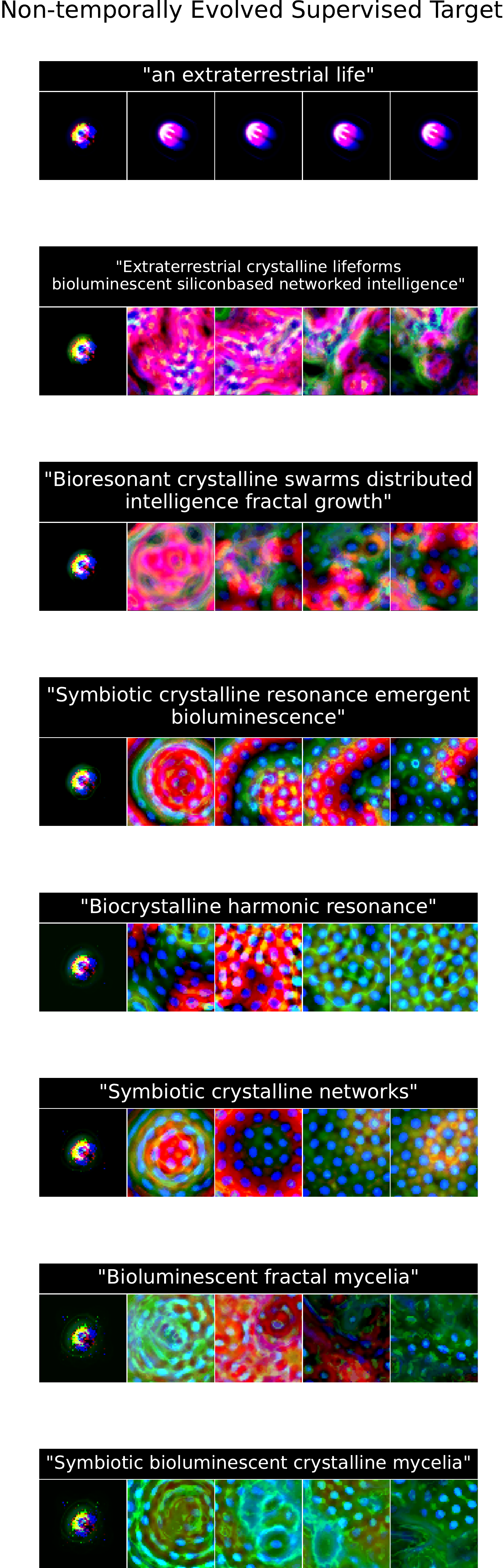}
    \caption{\textbf{Evolved simulations in Lenia using \MethodName{} with EST}. Starting with \texttt{an extraterrestrial life}, Gemma-3 proposes a new target prompt at each iteration which is then optimised using a single-target supervised loss. Each row visualises the discovered simulation for that iteration.}
    \label{EST}
\end{figure}

\paragraph{Diverse generations.} We find that EST leads to more novelty across iterations, since at each iteration Gemma-3 is not explicitly constrained by the evolutionary history---we pass in only the most recent target prompt in the instruction. An example run is shown in \Cref{EST} for initial prompt \texttt{an extraterrestrial life}, demonstrating the breadth of diversity between iterations of the algorithm---the system is seen to jump between distinct morphological patterns between iterations. 

\begin{table}[hbtp]
  \centering
    \resizebox{\linewidth}{!}{%
      \begin{tabular}{@{}lrr@{}}
        \toprule
        Initial Prompt           & OE score (mean $\pm$ std) & $\Delta$OE score\\
        \midrule
        a pepperoni pizza        & 0.058 $\pm$ 0.006               & 0.0168           \\
        a slime mould            & 0.058 $\pm$ 0.005               & -0.002           \\
        a flower                 & 0.045 $\pm$ 0.004               & 0.012            \\
        the garden of eden       & 0.049 $\pm$ 0.002               & 0.001            \\
        an extraterrestrial life & 0.0575 $\pm$ 0.010              & 0.027            \\
        a monkey                 & 0.052 $\pm$ 0.006               & 0.016            \\
        a caterpillar            & 0.054 $\pm$ 0.007               & 0.023            \\
        a microbe                & 0.044 $\pm$ 0.005               & -0.012           \\
        a fungus                 & 0.048 $\pm$ 0.004               & 0.004            \\
        \midrule
        MEAN                     & 0.052 $\pm$ 0.002               & 0.009 $\pm$ 0.004  \\
        \bottomrule
      \end{tabular}%
    }
    \caption{\textbf{OE scores from EST experiments}. Left column shows mean and standard deviation of OE score calculated for each resulting simulation from 8 \MethodName{} iterations. $\Delta$OE score is the difference between OE scores of the final and initial simulation for each run.}
    \label{tab:high_level_summary_sub2}
\end{table}

\paragraph{OE scores.} OE scores of simulations generated by EST are displayed in \Cref{tab:high_level_summary_sub2}. These scores measure novelty \textit{within} each simulation. The mean difference between the open-endedness of the final and initial simulations is more than two standard deviations above 0 at $0.009 \pm 0.004$, showing that \MethodName{} brings an increase in diversity over ASAL. Also, we find that the mean OE score across runs is $0.052 \text { vs ETT's  }0.049$ (\Cref{tab:high_level_summary_sub1}), more than one standard deviation higher than ETT.

\paragraph{Limited continuity.} Despite the diversity across iterations, we find that thematic coherence is harder to observe in the frames and it is harder to see influence of the past prompts in a particular iteration. For example, the purple glider in the first iteration of \Cref{EST} disappears by the second iteration, and later iterations bear no resemblance to earlier ones. This may be desirable if the aim is to generate as novel an evolutionary trajectory as possible, but may be prohibitive if the aim is to build coherent, linked stages of evolution. 

\paragraph{Occasional vanishing.} Visually, vanishing activations exist slightly in EST. However, we find that the independence from past prompts encourages perturbation from the best set of parameters converged on in previous iterations, and future iterations show reappearing activations. This method is better for continuous open-endedness.

\subsection{Evolved Temporal Targets}

\begin{table}[htbp]
  \centering
    \resizebox{\linewidth}{!}{%
      \begin{tabular}{@{}lrr@{}}
        \toprule
        Initial Prompt           & OE score (mean $\pm$ std) & $\Delta$OE score\\
        \midrule
        a pepperoni pizza        & 0.057 $\pm$ 0.010               & 0.018            \\
        a slime mould            & 0.049 $\pm$ 0.002               & 0.002            \\
        a flower                 & 0.044 $\pm$ 0.006               & 0.004            \\
        the garden of eden       & 0.057 $\pm$ 0.006               & -0.002           \\
        an extraterrestrial life & 0.037 $\pm$ 0.001               & -0.0001          \\
        a monkey                 & 0.046 $\pm$ 0.005               & 0.005            \\
        a caterpillar            & 0.049 $\pm$ 0.006               & 0.014            \\
        a microbe                & 0.051 $\pm$ 0.002               & -0.003           \\
        a fungus                 & 0.047 $\pm$ 0.007               & -0.020           \\
        \midrule
        MEAN                     & 0.049 $\pm$ 0.002               & 0.002 $\pm$ 0.004  \\
        \bottomrule
      \end{tabular}
      }
    \caption{\textbf{OE scores from ETT experiments}. Left column shows mean and standard deviation of OE score calculated for each resulting simulation from 8 \MethodName{} iterations. $\Delta$OE score is the difference between OE scores of the final and initial simulation for each run.}
    \label{tab:high_level_summary_sub1}
\end{table}

\begin{figure}
    \centering
    \includegraphics[width=0.71\linewidth]{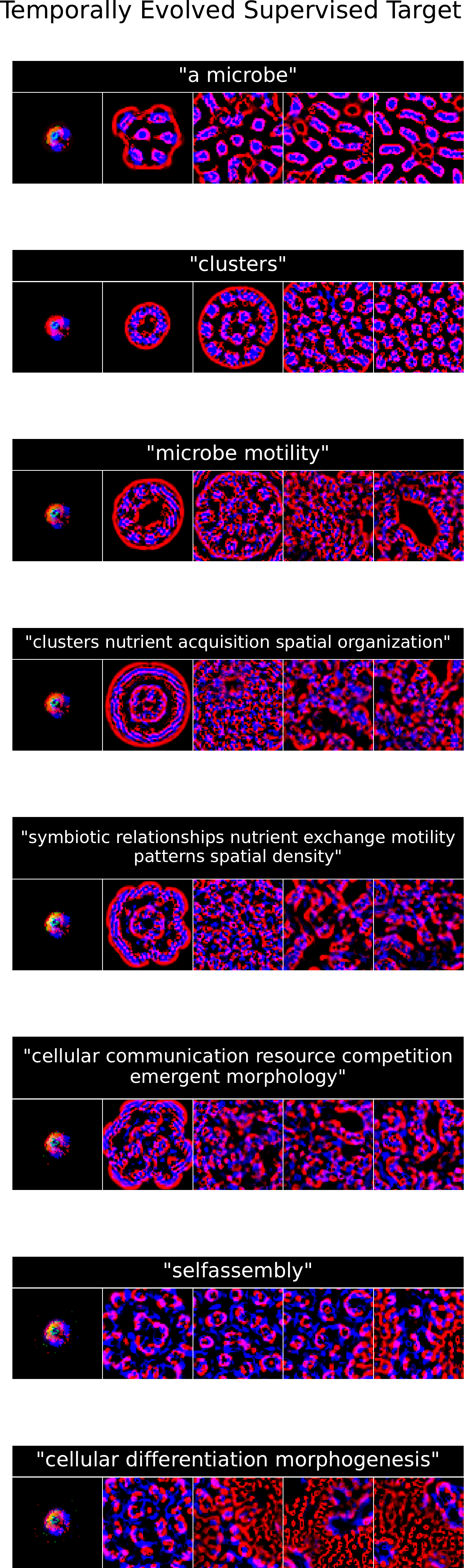}
    \caption{\textbf{Evolved simulations in Lenia using \MethodName{} with ETT.} Starting with \texttt{a microbe}, each row displays the optimised rollout from each iteration of \MethodName{}. Each simulation is optimised for the preceding Gemma-3 generated list of prompts.}
    \label{ETT}
\end{figure}

\paragraph{Evolutionary continuity.} Since, in ETT, Gemma-3 receives a full list of previous prompts, this generally yields prompt sequences which build fairly coherently from each other---for example, evolving from \texttt{a microbe} to \texttt{clusters} to \texttt{microbe motility} in \Cref{ETT}. The temporal passing of these lends to the interpretability of this method as we see elements of previous prompts in a given iteration - for example, in the third iteration of \Cref{ETT}, we see a microbe like structure grow to discrete clusters and then move around.

\paragraph{OE scores.} Results for ETT with the 9 initial prompts are shown in \Cref{tab:high_level_summary_sub1}. The difference OE score has a mean of $0.002 \pm 0.004$ indicating that, within the error, we remain around baseline. While it does not improve, this method can continue to produce iterations that have a similar amount of novelty within iterations to ASAL. 

\paragraph{Local minima.} With ETT, we find that the algorithm sometimes converges to local minima after a few iterations and subsequent prompts are not sufficiently different to push the simulation back into novel behaviours e.g. for \texttt{a caterpillar}.
By passing previous prompts into the instruction, Gemma-3 is likely biased to produce similar next prompts. For example, for the initial prompt \texttt{a flower}, Gemma-3 repeated \texttt{fragment} in the last four iterations.
Also, with every iteration, since the new prompt is appended to the list of previous prompts, it has less influence, which causes successive simulations to look similar. The softmax coefficient can encourage the model to fit each distinct prompt more closely. A higher Gemma-3 temperature could also reduce the frequency of repeating next prompts.

\paragraph{Vanishing activations.} In some simulations, activations vanish and don't re-appear.
FlowLenia \citep{plantec2023flowleniaopenendedevolutioncellular} is an extension of Lenia which avoids vanishing activations using mass conservation, but this creates a trade-off with the expressivity of the substrate and preliminary experiments did not show interesting behaviour with \MethodName{}.

\subsection{Trees of Life}

\begin{figure*}[htbp]
    \centering
    \includegraphics[width=0.65\linewidth]{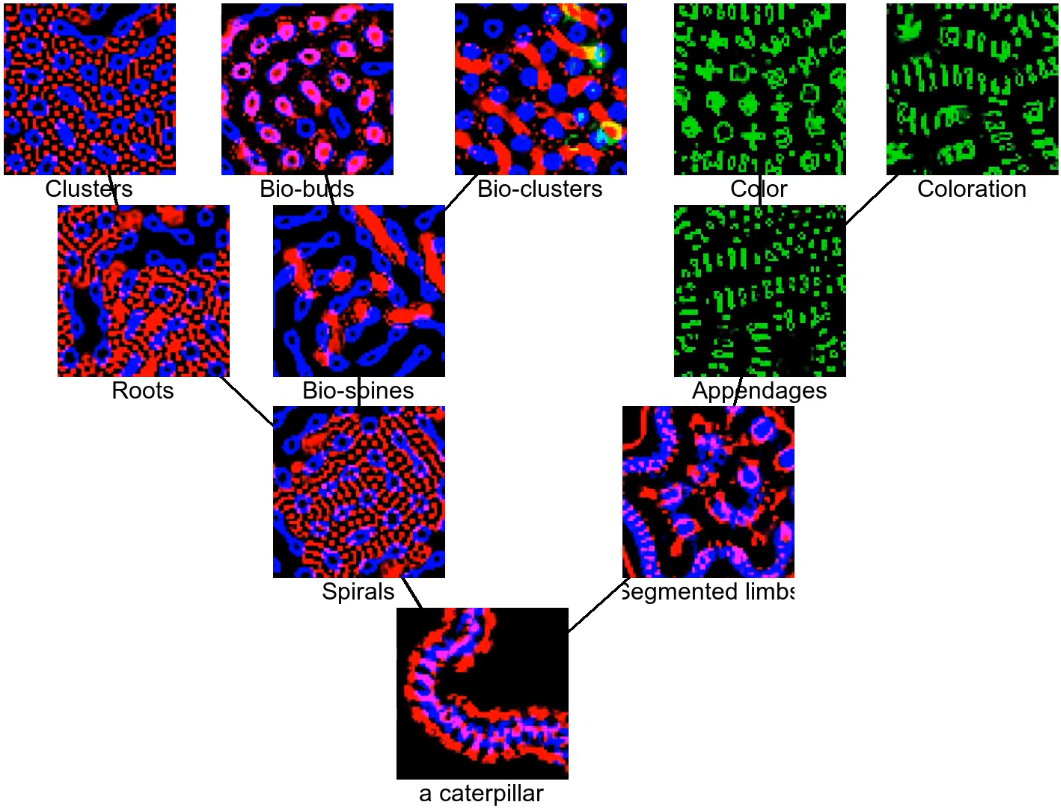}
    \caption{\textbf{Phylogenetic tree in Lenia.} We use initial prompt \texttt{a caterpillar} and resample the next target prompt from Gemma-3 multiple times to generate alternative targets. We display the final frame from each evolved simulation. }
    \label{tree1}
\end{figure*}


\begin{figure*}[htbp!]
    \centering
    \includegraphics[width=0.8\textwidth]{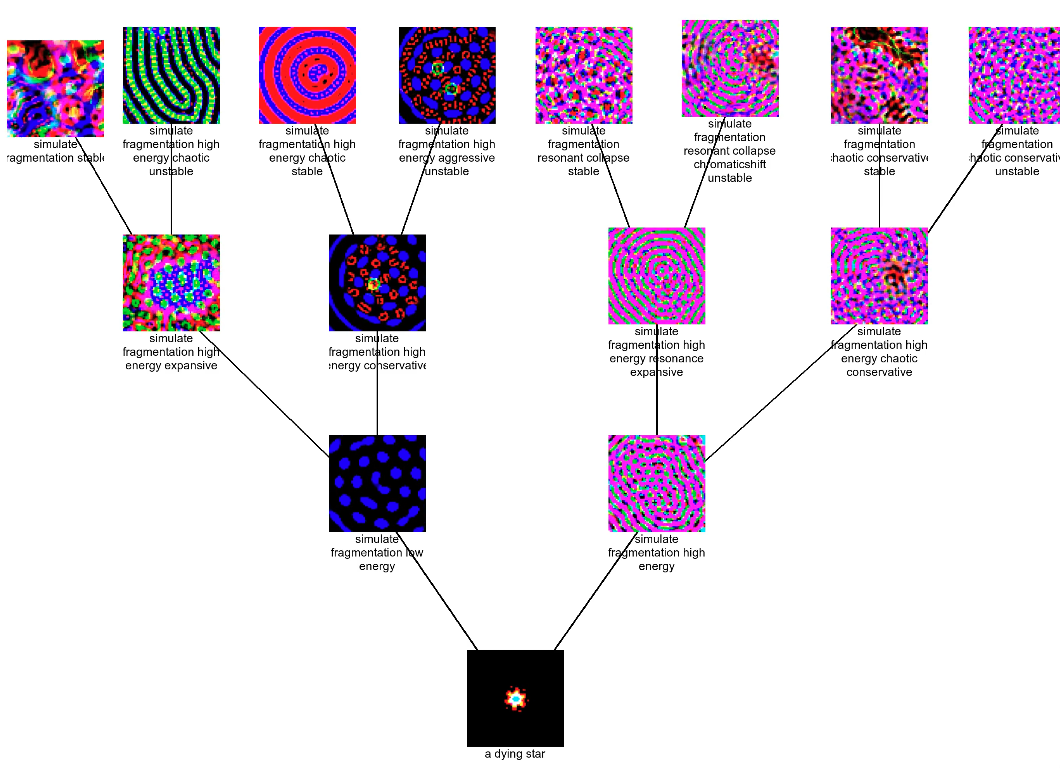}
    \caption{\textbf{Phylogenetic tree in Lenia using environmental inputs.} We use initial prompt \texttt{a dying star} and display the final frame from each evolved simulation. The environmental prompts have been appended to the end of each Gemma-3 prompt. The first split has the same Gemma-3 prompt with ``high energy'' and ``low energy'' applied to the nodes respectively. This creates quite different iterations as seen in the final frames.
    }
    \label{fig:tree2}
\end{figure*}

We generate phylogenetic trees in Lenia by resampling alternative next target prompts in ETT, using a higher temperature for Gemma-3.

Phylogenetic trees provide a qualitative view of the branching capacity of \MethodName{}. By resampling prompts, we can visualise parallel evolutionary trajectories rather than a single chain, illustrating both the method’s exploratory potential and the role of environmental pressures in shaping diverse outcomes.

Such a ``tree of life'' starting from \texttt{a caterpillar} is shown in \Cref{tree1}; after only a few iterations of \MethodName{}, we observe high diversity between branches.



\paragraph{Environmental pressure.} \label{sec:env_pressure} We take inspiration from physical environments and add additional pressures in the form of prompts we append at the end of the instruction prompt passed to Gemma-3. An example starting with \texttt{a dying star} is shown in \Cref{fig:tree2}, where we fork the tree with pairs [high energy, low energy], [expansive, conservative], [stable, unstable]---generated by GPT-4---applied to each layer respectively. These perturbations promote greater visual diversity. However, the additional prompting of GPT-4 for these environmental pressure prompts makes this method computationally expensive.

\subsection{Limitations and further work}

\paragraph{Biological bias.} Our experiments show a bias towards vocabulary related to biology or cellular automata. For example, the Gemma-3 suggestion for `a pepperoni pizza', even for ETT, effectively ignores this and suggests `pulsating bioluminescence'. We conjecture that the Gemma-3 model is trained on data that relates Lenia simulations with cellular automata or biological information. Larger FMs, or FMs fine-tuned for our purposes, could address this. 

\paragraph{More expressive substrates.} Lenia exhibits radial symmetry that favours circular patterns due to its Gaussian growth function \citep{wang2019lenia}. To increase expressivity we could use more substrates and especially multi-dimensional neural cellular automata (NCA). We tried testing this but due to memory constraints we could only use 1D-NCA which was far less novel than Lenia. 
 
\paragraph{Low resolution images.} CLIP inputs are images, a model that can take multiple frames could capture more meaningful embeddings for our task, like VideoCLIP \citep{xu2021videoclipcontrastivepretrainingzeroshot} or CLIP4CLIP \citep{Luo2021CLIP4Clip}.

\paragraph{Running more inner iterations of ASAL.} We find using more inner iterations often leads to less vanishing simulations and better fits. For example, we test 'the caterpillar' as an initial prompt on 4000 inner iterations, instead of 2000, and it produced a shape far more reminiscent of a caterpillar.

\paragraph{Dynamic OE scores.} We could dynamically calculate an OE score across iterations of a given chain to decide when to stop iterating, decreasing cost and prioritising novel routes.

\paragraph{Additional quantitative benchmarks.} While our analysis primarily relies on OE scores, inspired by \citep{kumar2024automating}, future work could incorporate additional quantitative benchmarks from recent literature (e.g. measures proposed in \citep{michel2025exploringflowleniauniversescuriositydriven}) to provide complementary perspectives. Comparing ASAL++ trajectories against diverse statistical measures—beyond OE scores—would strengthen robustness claims and allow alignment with continuous CA research.

\paragraph{Quality-diversity search.} An interesting direction would be to use a quality-diversity algorithm such as MAP-Elites \citep{mouret2015illuminating} and use \MethodName{} to generate a vast archive of diverse simulations.

\paragraph{Multi-agent interactions.} It would be interesting to promote diversity among patterns within a simulation---for example, FlowLenia admits multiple interacting organisms with different rule parameters. One could use a FM such as Gemma-3 to analyse a simulation video to classify distinct lifeforms, and use this to push diversity within a single simulation.

\section{Related work}
\label{sec:Related work}
The quest to automate the discovery of ALife forms is not new. LeniaBreeder \citep{faldor2024leniabreeder} uses quality-diversity (QD) algorithms \citep{pugh2016quality} to evolve self-organising patterns in Lenia. LeniaBreeder uses both MAP-Elites \citep{mouret2015illuminating}, which relies on a pre-defined discretisation of the search spaces into ecological niches, as well as AURORA \citep{grillottidiscovering}, an unsupervised QD algorithm. While AURORA can dynamically redefine niches, the possible evolutionary trajectories are limited by the chosen statistical measures in the algorithm. In addition, there is no way for humans to qualitatively describe the outcome they want. In contrast, our method leverages the representation abilities of FMs from their large-scale training to drive the evolution of artificial life in much more open-ended ways.

Other attempts have been made to build large-scale open-ended simulations. One such is JaxLife \citep{lu2024jaxlife}, an agentic open-ended simulator which aims to facilitate socio-cultural behaviours as opposed to cellular behaviours simulated by most ALife substrates.

In addition, \citet{chan2023towards} explores open-ended evolution in a substrate based on Lenia through large-scale intrinsic simulations, aiming to replicate the emergence of open-ended self-organizing patterns. However, their method lacks a way to actively and automatically guide the discovery process to allow for open-ended evolution, relying on specific design choices and tending to converge to optimal solutions.


\section{Conclusion}
\label{sec:Conclusion}
We presented \MethodName{}, a novel framework for artificial life simulation exploration that displays open-ended characteristics through the use of foundation models. Our first strategy, \textit{Evolved Supervised Targets}, promotes higher visual novelty and open-ended behaviours by evolving toward a single new target per iteration. In contrast, our second strategy, \textit{Evolved Temporal Targets}, fosters stable trajectories and interpretability by optimising for the entire temporal history of generated prompts.

While our method suggests new directions towards open-ended ALife discovery, it is limited by the biases of foundation models, substrate constraints and computational costs. Future work could build on ours by using more expressive ALife substrates, or combining the power of FM-guided evolution with more sophisticated search techniques such as quality-diversity.

\paragraph{Acknowledgements.} We thank Laura Ruis for her supervision and guidance, Akarsh Kumar for his feedback, and Catherine Bacon for editing.

\newpage
\footnotesize
\bibliographystyle{apalike}




\end{document}